\documentclass{article}

\usepackage{arxiv}

\usepackage[utf8]{inputenc} 
\usepackage[T1]{fontenc}    
\usepackage{hyperref}       
\usepackage{url}            
\usepackage{booktabs}       
\usepackage{amsfonts}       
\usepackage{nicefrac}       
\usepackage{microtype}      
\usepackage{lipsum}
\usepackage{fancyhdr}       
\usepackage{graphicx}       
\graphicspath{{media/}}     
\usepackage{comment}

\title{Question Answering on Patient Medical Records with Private Fine-Tuned LLMs

}

\author{
  Sara Kothari\\
  Department of Computer Science\\
  Stanford University \\
  \texttt{sarako@stanford.edu} \\
   \And
  Ayush Gupta \\
  Genloop Labs, Inc. \\
  Delaware, USA \\
  \texttt{founder@genloop.ai} \\
}

\begin{document}
\maketitle

\begin{abstract}

Healthcare systems continuously generate vast amounts of electronic health records (EHRs), commonly stored in the Fast Healthcare Interoperability Resources (FHIR) standard. Despite the wealth of information in these records, their complexity and volume make it difficult for users to retrieve and interpret crucial health insights. Recent advances in Large Language Models (LLMs) offer a solution, enabling semantic question answering (QA) over medical data, allowing users to interact with their health records more effectively. However, ensuring privacy and compliance requires edge and private deployments of LLMs.

This paper proposes a novel approach to semantic QA over EHRs by first identifying the most relevant FHIR resources for a user query (Task1) and subsequently answering the query based on these resources (Task2). We explore the performance of privately hosted, fine-tuned LLMs, evaluating them against benchmark models such as GPT-4 and GPT-4o. Our results demonstrate that fine-tuned LLMs, while 250x smaller in size, outperform GPT-4 family models by 0.55\% in F1 score on Task1 and 42\% on Meteor Task in Task2. Additionally, we examine advanced aspects of LLM usage, including sequential fine-tuning, model self-evaluation (narcissistic evaluation), and the impact of training data size on performance.
The models and datasets are available here: \href{Hugging Face Link}{https://huggingface.co/genloop}

\end{abstract}


\section{Introduction}

The Fast Healthcare Interoperability Resources (FHIR) \cite{hl7} standard, developed by HL7, provides a consistent framework for exchanging electronic health records (EHRs) across healthcare systems, enabling improved interoperability. FHIR’s structured, machine-readable format supports the seamless transfer of complex healthcare data, making it central to modern health information exchange. 

Recent mandates requiring the use of HL7 FHIR Release 4 to support API-enabled services reflect a shift toward greater data transparency and patient autonomy\cite{cmsInteroperabilityPatient}. Regulatory changes, particularly the Anti-Information Blocking provisions of the 21st Century Cures Act \cite{united_states_national_archives_and_records_administration_office_of_the_federal_register_act_2016}, also emphasize the need for patient-centric access to health data. Platforms like Apple Health now integrate FHIR data, giving patients unprecedented access to their medical records and creating new opportunities to transform complex data into actionable health insights. \cite{appleHealthcareHealth}

Despite these advances in interoperability, patients often face significant challenges when interacting with their own health data. Complex medical terminology, language barriers, and limited health literacy make it difficult for many individuals to understand their conditions, diagnoses, or treatment plans \cite{al_shamsi_implications_2020}. These challenges are not just inconveniences—they can lead to delays in care, misunderstandings of critical information, and ultimately, poorer health outcomes.

Large Language Models (LLMs) \cite{arxivLargeLanguage} provide a paradigm for human readability of their health records. LLMs, characterized by their ability to process vast amounts of unstructured data and generate human-like text, are transforming the landscape of healthcare informatics. Their inherent capacity to perform natural language understanding (NLU) and generation enables them to extract, summarize, and interpret complex medical information. 

In the context of this paper, LLMs serve as a bridge between raw FHIR (Fast Healthcare Interoperability Resources) data and end-users, translating intricate clinical terminology into plain language that patients and healthcare providers can easily comprehend. By leveraging Large Language Models (LLMs), healthcare providers have the opportunity to simplify and personalize health data. Such systems can empower patients, making their medical information more accessible and helping them better understand their health, which in turn may enhance patient engagement and adherence to treatment plans. This approach also has the potential to reduce healthcare inefficiencies by offering quicker, more intuitive access to relevant medical records.

However, the application of LLMs for such tasks faces significant barriers, particularly with respect to data privacy and security. Sharing personal health information (PHI) with cloud-hosted, general-purpose LLMs risks violating key privacy regulations, such as HIPAA \cite{rights_ocr_health_2021}, the California Consumer Privacy Act (CCPA)\cite{noauthor_california_nodate}, and the Biometric Information Privacy Act (BIPA, Illinois) \cite{noauthor_biometric_2021}. In the healthcare domain, ensuring the confidentiality of sensitive patient data is paramount. For this reason, edge-deployed, privately hosted LLMs present a more viable solution, allowing healthcare systems to maintain control over patient data while still benefiting from the powerful capabilities of LLMs.

Thanks to open-source models like Meta’s LLaMA series \cite{touvron_llama_2023, grattafiori2024llama3herdmodels} and Mistral’s open source model series \cite{jiang2023mistral7b}, it is now feasible to self-host LLMs for specific applications. However, these smaller models often fall short in the accuracy required for sophisticated healthcare tasks such as semantic question answering. Fine-tuning with techniques like LoRA \cite{houlsby2019parameter, hu_lora_2021, dettmers_qlora_2023} emerges as a critical step to address this gap with computational efficiency. By customizing smaller, domain-specific models to the nuances of the target task, fine-tuning enhances both the accuracy and performance of these models. This adaptation not only makes them suitable for demanding healthcare applications but also ensures they remain efficient and privacy-compliant, aligning with the requirements of edge deployment in sensitive domains like healthcare.

In this paper, we break down the task of semantic question answering over medical records into two stages: (1) retrieving the most relevant FHIR resources given a user’s medical query, and (2) answering the query based on the retrieved resources. We fine-tune smaller, open-source models for each stage, ensuring that they are well-suited to the unique challenges of medical data processing. To facilitate this process, we generate synthetic patient data and utilize larger general-purpose models to create task-specific synthetic datasets (data collection). These datasets are then refined (data refinement), followed by training multiple models to identify the best-performing configurations (training), and finally, evaluating their performance against established benchmarks (evaluation). The resulting models are deployed in a privacy-compliant setup.

Our experiments reveal that fine-tuned smaller models significantly outperform larger general models like GPT-4, particularly in terms of accuracy, efficiency, and privacy compliance. This work demonstrates that edge-deployed, fine-tuned LLMs can offer a practical solution for healthcare providers seeking to implement patient-centric semantic question answering without sacrificing data privacy.

Beyond the core tasks, this study explores several important aspects of LLM behavior in healthcare contexts:

\begin{enumerate}
\item  The impact of sequential fine-tuning on multi-task performance.
\item  The tendency of LLMs to exhibit narcissistic behavior by favoring their own outputs.
\item  The influence of dataset size on fine-tuning effectiveness, with implications for resource-constrained environments.
\end{enumerate}

We discuss these results in Section 5 (Results and Discussion) \ref{sec:results} and highlight key insights for future research in Section 6 (Conclusion, Limitations, and Future Work) \ref{sec:conclusion}.

\section{Related Work}
\label{sec:related_work}
The application of Large Language Models (LLMs) to patient medical data processing has garnered significant attention in recent research. Notably, \cite{li_fhir-gpt_2023} demonstrated the efficacy of leveraging LLMs to convert clinical texts into FHIR resources with an impressive accuracy rate of over 90 percent, thereby streamlining the processing and calibration of healthcare data and enhancing interoperability. \cite{sudarshan_agentic_2024} introduced a multi-agent workflow utilizing the Reflexion framework, which employed ICD-10 codes from medical reports to generate patient-friendly letters with an accuracy rate of 94.94\%. These studies did not involve direct querying of Electronic Medical Records (EMRs).\\
Recently, \cite{schmiedmayer_llm_2024} introduced "LLM on FHIR", an open-source mobile application that leverages GPT-4 to allow users to interact with their health records, demonstrating impressive accuracy and relevance in delivering comprehensible health information to patients. \cite{brockman_supahot_2024} developed on top of \cite{schmiedmayer_llm_2024} to replace GPT-4 with fine-tuned LLMs. They divided the FHIR querying process into three tasks: filtering, summarization, and response. They compared against Meditron\cite{meditron}, a family of medical-adapted Llama-2 models, and Llama 2 Base models. Their approach was shown performing better than Llama 2 Base models but underperforming Meditron.

\section{Approach}
\label{approach}
To approach this task, we break query processing in 2 stages, similar to how retrieval augmented generation \cite{gao2024retrievalaugmentedgenerationlargelanguage, wu2024retrievalaugmentedgenerationnaturallanguage}, is performed.
\begin{enumerate}
    \item Task 1: Identifying the FHIR resources from the patient’s medical record that are relevant to a given natural language patient query. Each patient will have numerous FHIR resources in their patient record, only some of which are relevant to the patient query. We formulate the problem as a binary classification problem i.e. given a query q, and a FHIR resource r, the problem is setup as
    F(q, r) = I\{0, 1\}
    \item Task 2: Answering the medical query of the patient using the FHIR resources that were identified as relevant to the query i.e. generating the response based on (query, list of relevant resources) pairs.
\end{enumerate}

\begin{figure}
  \centering
  \includegraphics[width=\textwidth]{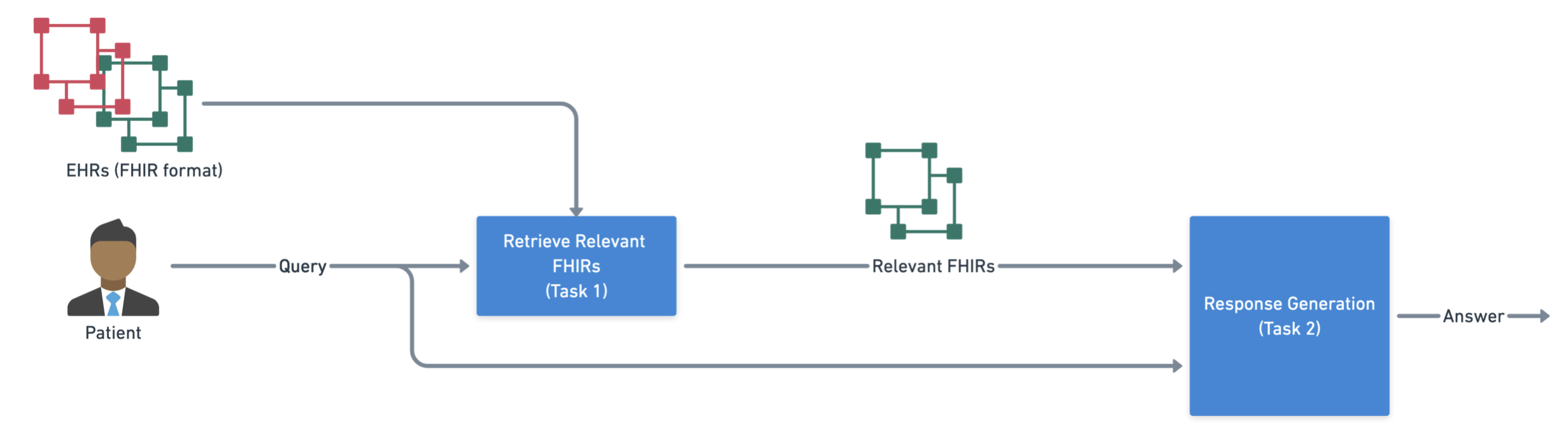}
  \caption{Complete Approach Diagram}
  \label{fig:fig1}
\end{figure}

Figure 1 \ref{fig:fig1} outlines the approach. The main inputs are the user query and the EHR records in the FHIR format. The EHR records can be both relevant or irrelevant when received. The Task 1 classifies relevant FHIRs and help us pick the necessary records to generate the answer for the user's medical question in Task 2. 

LLMs are the intelligence modules for Task 1 and Task 2. In our approach, we fine-tune the top-performing models available at the time of experimentation (Llama-3.1-8B, and Mistral-NeMo) for each of these tasks. We also compare the results switching them with GPT-4 (SOTA model at the time), and Meditron-7b \cite{meditron} (SOTA medical domain model at the time). More details on the experiments, including dataset generation, are covered in Section 4 (Experiment Details). The Results are discussed in Section 5 (Results and Discussion)

\section{Experiment Details}
\label{sec:modelling}
\subsection{Dataset}
We employed Synthea \cite{walonoski_synthea_2018}, an open-source synthetic patient generator, to produce HL7 FHIR-formatted medical records (JSON files). we started with generating data of 50 patients for Task 1 and an additional 450 patients for Task 2. Each file was processed to retain only patient-relevant resources, excluding entries such as "SupplyDelivery". Only the following resource types were included: {"Procedure'', "Medication", "MedicationRequest", "Encounter", "ImagingStudy", "Immunization", "Device", "CarePlan", "ExplanationOfBenefit", "AllergyIntolerance", "Observation", "Condition", "DiagnosticReport"} \

We ran a custom script to filter and retain only those segments of each FHIR resource that are likely to be relevant to patient queries. The script eliminated non-essential details and sections from which meaningful medical information could not be extracted. This pre-processing step was designed to eliminate irrelevant or redundant information. 

\begin{figure}
  \centering
  \includegraphics[width=\textwidth]{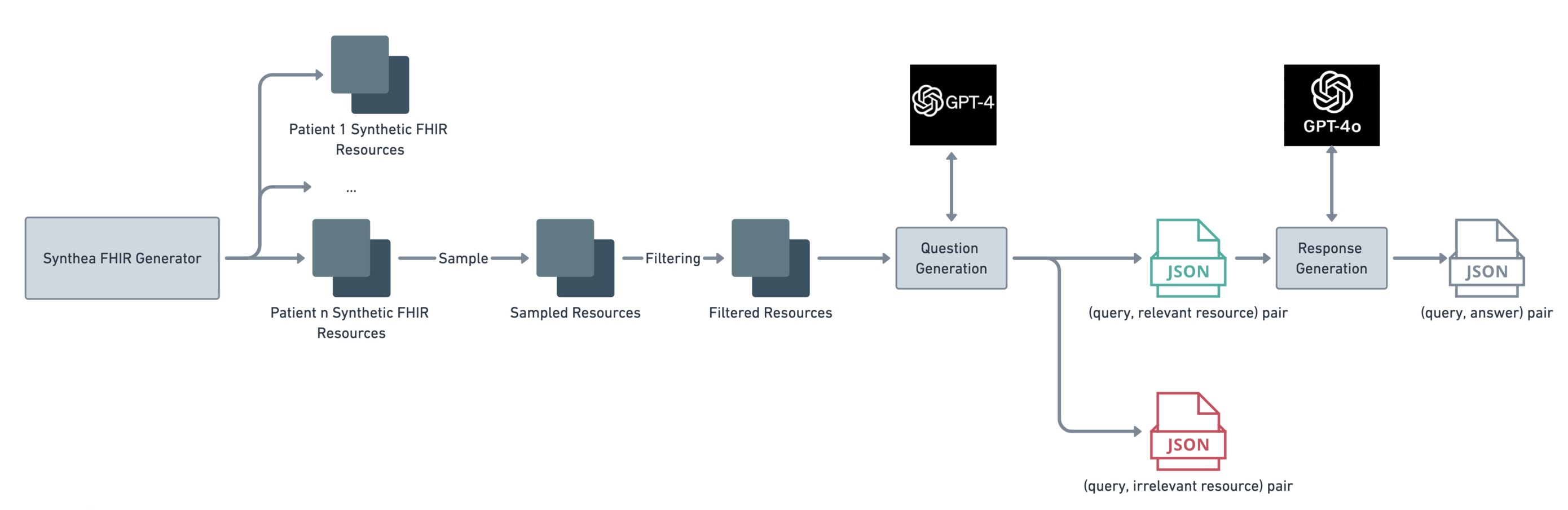}
  \caption{Data Preparation Flow}
  \label{fig:fig2}
\end{figure}

Figure 2 \ref{fig:fig2} shows the complete data preparation flow. More details on creating the training pairs for Task 1 LLM fine-tuning and Task 2 LLM fine-tuning are covered in the following sections.

\subsubsection{Task 1 Data Preparation}
The following additional steps were followed for Task 1 Data Preparation
\begin{enumerate}
    \item \textbf{Data Selection}: We randomly selected a batch of 10 FHIR resources from each patient record.
    \item \textbf{Query Generation}: GPT-4 was employed to generate a natural language query based on one or more of the 10 selected FHIR resources. We utilized prompt engineering to ensure the generation of realistic, simple, and non-technical queries. The specific prompt used has been provided in Appendix \ref{sec:appendix-prompt}.
    \item \textbf{Labeling}: GPT-4 labeled the resources as "relevant" if they were used to generate the query and "irrelevant" if they were not.
    \item \textbf{Dataset Creation}: This process was repeated 10 times per patient, resulting in a total of 5,000 JSON files containing queries, patient IDs, relevance label, and resource. For LLM fine-tuning, the inputs are the query and resource, and the output is relevance label. We used a 95-5 training-test split for the dataset.

\end{enumerate}

\subsubsection{Task 2 Data Preparation}
For Task 2, we generated 450 additional patient records and processed them through the same method as Task 1 Data Preparation. That gave us a total of 500 patient records, with the corresponding relevance label for each resource,. 
\begin{enumerate}
    \item \textbf{Answer Generation}:  For each query, we have relevant and irrelevant FHIR resources. For the task of answer generation, we are interested in only the relevant resources. Therefore, each input example for Task 2 is a pair of (query, [list of relevant resources]). Since we have 10 queries per patient record, we have 5,000 such examples. For each of them, GPT-4 was used to create a relevant answer. The generated answered were reviewed for accuracy along with their context. Finally, we had 5,000 examples of (query, [list of relevant resources], answer). The specific prompt used for GPT-4 based synthetic data generation has been provided in Appendix \ref{sec:appendix-prompt3}.
    \item \textbf{Dataset Creation}: We conducted a 98-2 train-test split, resulting in 4900 examples in the training dataset and 100 examples in the test dataset. We also randomly sub-sampled 500 examples from the 4900 example training dataset to create another smaller training dataset to determine the effect of training dataset size on model performance (details covered in following sections). 

\end{enumerate}

\subsection{Fine-tuning}
Fine-tuning \cite{houlsby2019parameter, hu_lora_2021, dettmers_qlora_2023} is a technique by which the original model is modified to adapt to the specific task of concern. Full fine-tuning of LLMs is a compute intensive process. For reference, a 8B parameter model will require around 100GB GPU VRAM, that is available with exclusive GPU machines like H200 costing more than \$10,000 per month on rent and hardly available to purchase.

Parameter-Efficient Fine-Tuning (PEFT) \cite{houlsby2019parameter} gives us a computationally efficient technique to perform fine-tuning with marginal performance difference. It adds additional trainable parameters to the LLM, that work in conjunction with the original pre-trained weights to give desired results. In our experiments, we use a variant of PEFT that provides even better memory efficiency called QLoRA \cite{dettmers_qlora_2023}. This technique allows to load the original model weights in 4-bit quantization reducing memory usage, while training additional weights through the PEFT approach. For our experiments, we used a LoRA rank of 16, and trained for 5 epochs. We conducted our fine-tuning and inference experiments on an NVIDIA A100 40GB GPU. More model training configs are available with the model cards on our HF page here: \href{Hugging Face Link}{https://huggingface.co/genloop}

\subsection{Experiments}
We additionally conduct the following experiments to ascertain how fine-tuned smaller LLMs perform against the general LLMs, and also empirically understand their behaviour:

\begin{enumerate}
    \item \textbf{Experiment 1 - Comparison of Base Models, Fine-tuned Models, Meditron, and GPT-4 Performance:} We compare the performance of pretrained models, models fine-tuned on specific tasks using custom datasets, and GPT-4 (and GPT-4o) to understand how finetuning improves task-specific performance and how these models stack up against a state-of-the-art model like GPT-4 or GPT-4o.

    \item \textbf{Experiment 2 - Analyzing the Effect of Training Dataset Size on Model Performance:}  In this experiment, we assess how varying the size of the training dataset impacts model performance.
     
    \item \textbf{Experiment 3 - Impact of Sequential Fine-tuning on Task Retention:} Through this experiment, we explore cross-task knowledge transfer. Specifically, how fine-tuning a model on a second task affects its ability to retain knowledge from the first task, assessing whether the model "forgets" how to perform the original task after being adapted to a new one, and vice versa. Moreover, it investigates whether being initially trained on a first task with knowledge overlap improves or hinders performance on the second task. 
    
    \item \textbf{Experiment 4 - Investigating Self-Preference and Self-Recognition Bias in LLM-as-a-Judge Evaluation:} This experiment analyzes potential biases in large language models when they act as evaluators, specifically focusing on whether the model exhibits a preference for its own outputs or recognizes and favors its own responses over others in an evaluation context.
\end{enumerate}

For \textbf{Task 1}, we conducted Experiment 1 and Experiment 3. 
\begin{enumerate}
    \item For \textbf{Experiment 1}, we fine-tuned Llama 3.1 Base, Llama 3.1 Instruct, Mistral NeMo Instruct, and Mistral NeMo Base using the 5,000-example dataset created for Task 1. We benchmarked these models against GPT-4 and Meditron-7b. The pretrained Llama 3.1 and Mistral Nemo Base models were our baseline.  
    \item For \textbf{Experiment 3}, we explored two approaches. First, we fine-tuned Llama 3.1 Base on Task 2 using a 4,900-example dataset, followed by additional fine-tuning on Task 1. In the second approach, we reversed the order by first fine-tuning Llama 3.1 Base on Task 1, then further fine-tuning the resulting model on Task 2. We did an experiment with an extended prompt for the second approach, in order to instruct the model on the classification requirements.
    
    We evaluated all these models on this binary classification task using F1, Recall, Precision, and Accuracy, placing greatest emphasis on F1. \\

\end{enumerate}

For \textbf{Task 2}, we conducted all four experiments. In all these experiments, we used the METEOR score \cite{banerjee2005meteor} as the evaluation metric. We chose METEOR as it incorporates stemming and synonymy matching, in addition to exact word matching, and aligns closely with human judgment at the sentence/segment level.
\begin{enumerate}
    \item For \textbf{Experiment 1}, we fine-tuned Llama 3.1 Base and Mistral NeMo Base using a dataset of 4,900 examples and benchmarked their performance against GPT-4o and Meditron 7B. 
    \item For \textbf{Experiment 2}, we fine-tuned Llama 3.1 Base and Mistral NeMo Base using a smaller dataset of 500 examples. We compared these models' performance to those fine-tuned in the first experiment (4,900 examples) to assess the effect of dataset size on model performance. 
    \item For \textbf{Experiment 3}, we conducted sequential finetuning on both Llama 3.1 Base and Mistral NeMo Base models, which were initially fine-tuned on Task 1. We further fine-tuned these models on Task 2 explore task transfer and task retention. Additionally, we fine-tuned the Llama 3.1 Base model, which had been first fine-tuned on Task 2 with 4,900 examples, on Task 1 using 5,000 examples, and then evaluated its performance on Task 2 to assess knowledge transfer across tasks. 
    \item For \textbf{Experiment 4}, we performed an LLM-as-a-judge evaluation involving the top two performing models from the previous tasks using Claude Sonnet and GPT-4o as judge LLMs. We initially asked GPT-4o to evaluate and select the better response between the two models without knowing which model generated which response. After this, we disclosed which response was generated by GPT-4 and asked GPT-4o to choose again. We then conducted the same blind evaluation process with Claude. 
    In both cases, the reference answer from the dataset was provided, and the evaluation criteria included the following:  
    \begin{enumerate}
        \item Relevance: Does the answer accurately address the patient query using the relevant FHIR resources?
        \item Groundedness: Is the answer based on factual, evidence-based information?
   \item Completeness: Does the answer cover all aspects of the patient query comprehensively?
   \item Quality: Is the answer well-written, clear, and useful for the patient?
   \item Conciseness: Is the answer succinct while still being informative?
   \item Closeness to Reference: How closely does the answer match the content and intent of the reference answer?

    \end{enumerate}

\end{enumerate}

\section{Results and Discussion}
\label{sec:results}
\subsection{Task 1 Results}

\subsubsection{Experiment 1} 
\begin{figure}
  \centering
  \includegraphics[width=\textwidth]{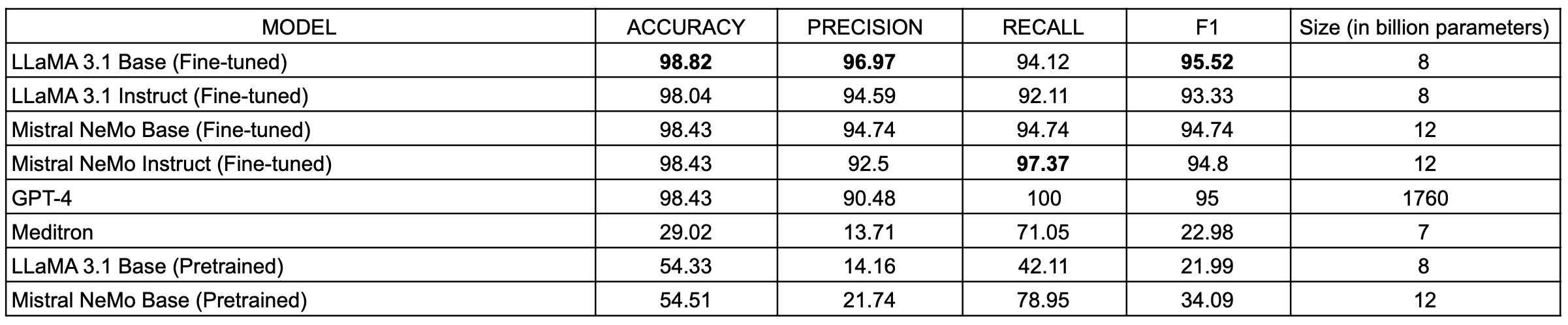}
  \caption{Task 1 Experiment 1 Results}
  \label{fig:fig3}
\end{figure}

The results for Task 1 Experiment 1 are shown in Figure 3 \ref{fig:fig3}. It can be seen that fine-tuning improves model performance significantly compared to the pre-trained baselines. LLaMA 3.1 Base (Fine-tuned) achieved the highest accuracy of 98.82\%, with strong performance in both precision (96.97\%) and recall (94.12\%), leading to the strongest F1 score of 95.52\%. This result highlights the model's effective learning and adaptation to Task 1 after fine-tuning, surpassing both GPT-4 and Meditron-7B. With 8B parameters, this model is approximately 0.45\% the size of GPT-4 (which has ~1760B parameters). On the other hand, LLaMA 3.1 Base (Pretrained), serving as a baseline, only achieved 54.33\% accuracy, while Mistral NeMo Base (Pretrained) had 54.51\% accuracy, reflecting their underperformance prior to fine-tuning. Another interesting point to note is that llama 3.1 base model has 8B parameters compared to Mistral NeMo with 12B parameters, but is able to give better results. 

\begin{figure}
  \centering
  \includegraphics[scale=0.5]{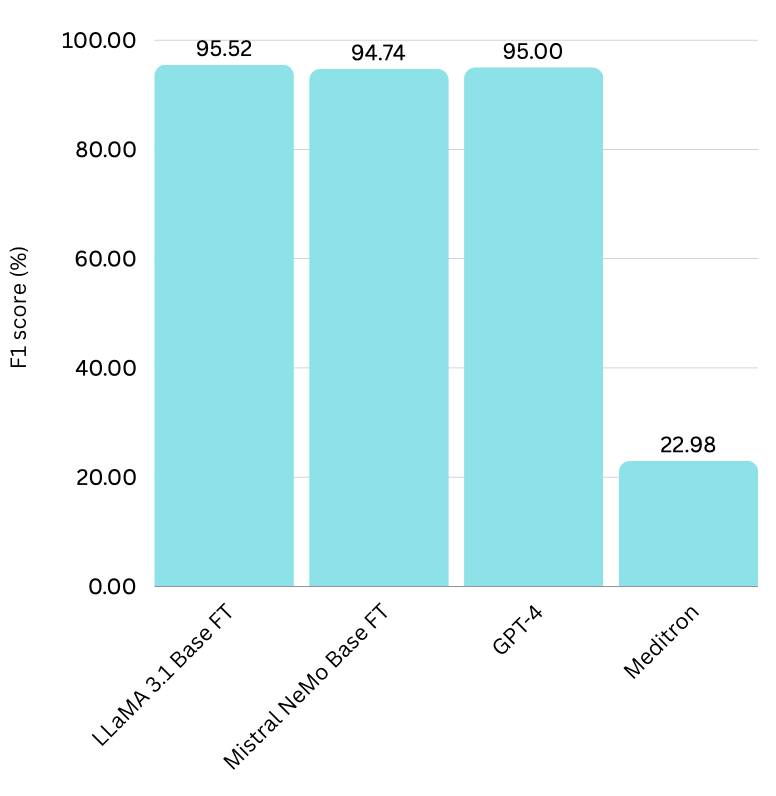}
  \caption{Task 1 Experiment 1 F1 Scores}
  \label{fig:fig4}
\end{figure}

Lastly, GPT-4, with an accuracy of 98.43\% and an F1 score of 95\%, was comparable to the fine-tuned models but lacked the same level of precision, highlighting the strength of task-specific fine-tuning. Meditron-7B, though a specialized medical model, underperformed significantly in this task with only 29.02\% accuracy. The models are further compared illustratively in Figure 4 \ref{fig:fig4}.

\subsubsection{Experiment 3} 
\begin{figure}
  \centering
  \includegraphics[width=\textwidth]{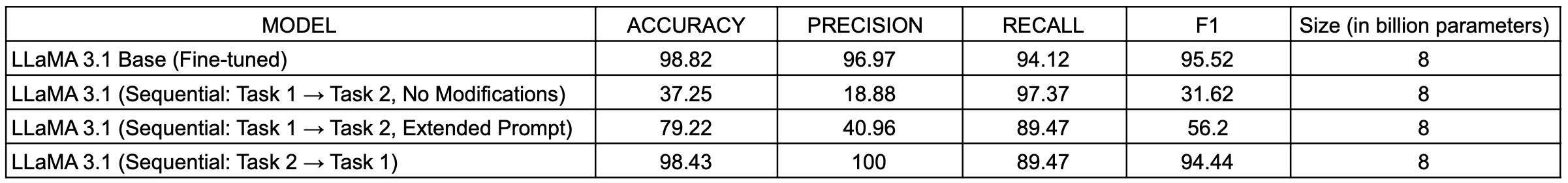}
  \caption{Task 1 Experiment 3 Results}
  \label{fig:fig5}
\end{figure}

The compiled results for this experiment of sequential training on Task 1 is presented in Figure 5 \ref{fig:fig5}. As expected, LLM trained on Task 1 and then on Task 2, loses its practice on Task 1 and gives inferior results. The F1 Score drops from 95.52 for a Llama 3.1 Base FT model to 31.62 for a Llama 3.1 Base model fine-tuned on Task 1 and then Task 2, as 67\% drop in F1. Understandably, Task 2 is more verbose than the classification problem in Task 1. So we conducted another experiment extending the prompt to instruct the classification into one word "relevant" or "irrelevant". This increased the F1 score to 56.62, a 78\% improvement. This helps understand that a fine-tuned model is still receptive to instructions in the prompt. Fine-tuning the LLM on Task 2 and Task 1 yields similar results as the Llama 3.1 Base directly fine-tuned on Task 1. 
 
\subsection{Task 2 Results}

\subsubsection{Experiment 1}
\begin{figure}
  \centering
  \includegraphics[width=0.6\textwidth]{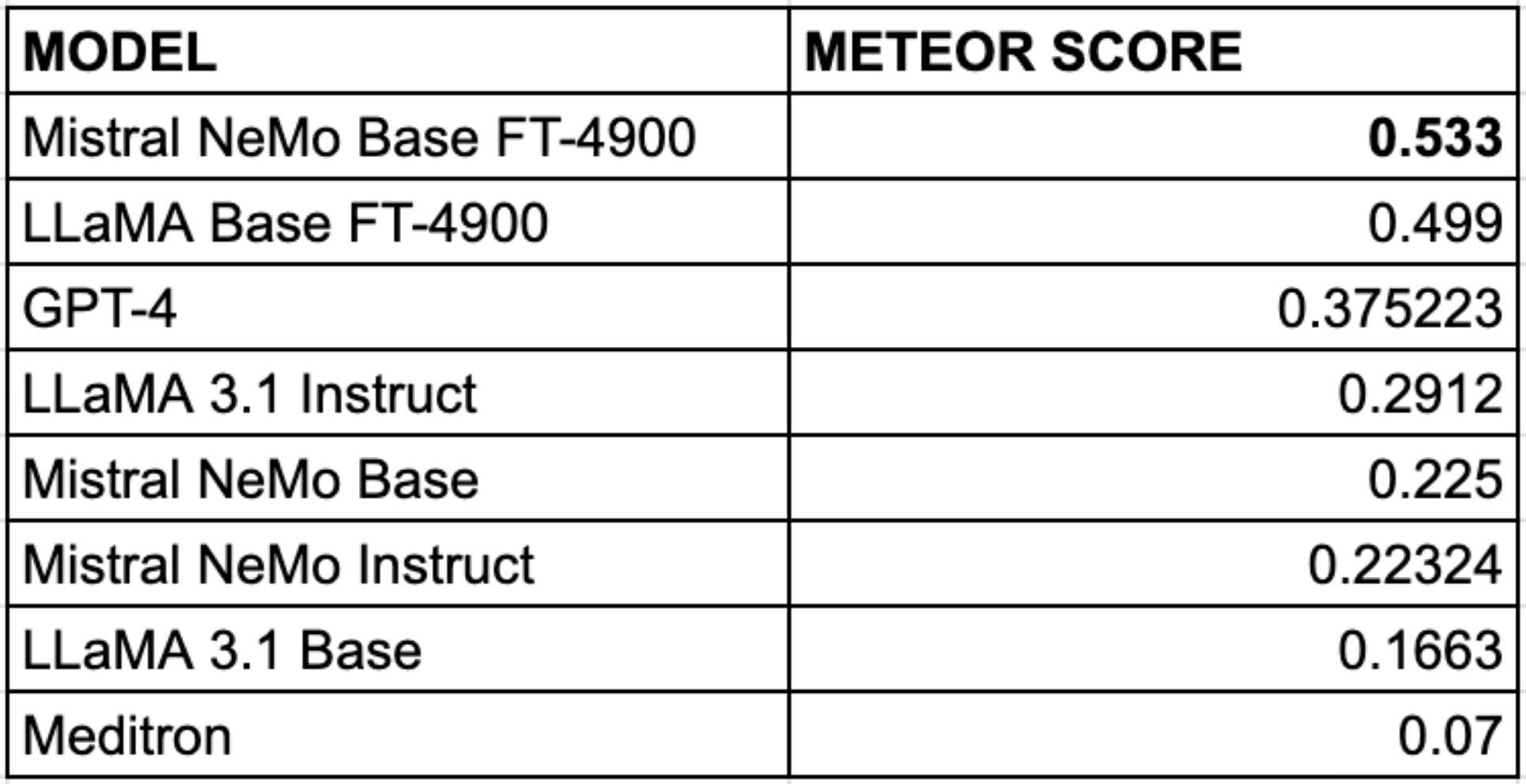}
  \caption{Task 2 Experiment 1 Results}
  \label{fig:fig6}
\end{figure}

Figure 6 \ref{fig:fig6} compiles the results from Experiment 1 on Task 2. We see that fine-tuned Llama 3.1 8B Base and Mistral NeMo 12B models surpass their base versions by 200\% and 136\%, respectively. Mistral NeMo Base fine-tuned on the 4,900 examples achieved the highest performance with a METEOR score of 0.5333, while GPT-4 scored lower at 0.375223. Meditron scored 0.07 on the same task.

\subsubsection{Experiment 2}
\begin{figure}
  \centering
  \includegraphics[width=0.6\textwidth]{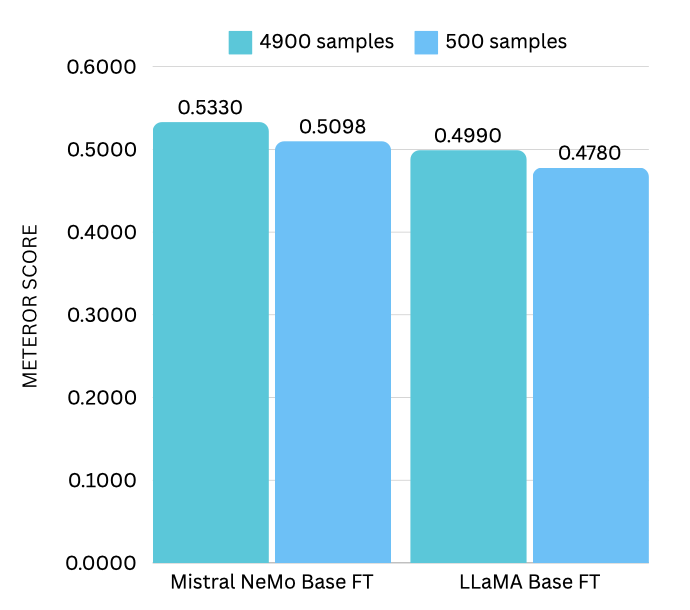}
  \caption{Task 2 Experiment 2 Results}
  \label{fig:fig7}
\end{figure}

For Experiment 2, we compared the Llama 3.1 and Mistral Nemo models trained on 4,900 examples against training on 500 examples. As can be seen in Figure 7 \ref{fig:fig7}, Mistral NeMo Base fine-tuned with 4900 examples outperformed its counterpart fine-tuned with 500 examples by 4.55\% in METEOR score. Similarly, LLaMA Base fine-tuned with 4900 examples showed a 4.39\% higher METEOR score compared to Mistral NeMo Base fine-tuned with 500 examples. We conclude that an increase in the training dataset size definitely improves model performance. However, a deeper study is required on more size variations and the limit to such improvement.

\subsubsection{Experiment 3}
\begin{figure}
  \centering
  \includegraphics[width=0.6\textwidth]{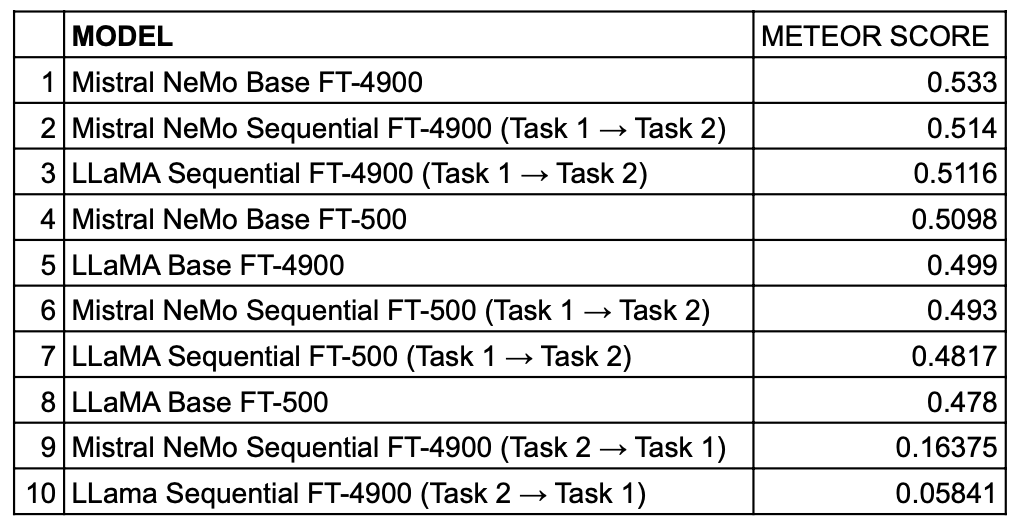}
  \caption{Task 2 Experiment 3 Results}
  \label{fig:fig8}
\end{figure}

The compiled results for this experiment of sequential training on Task 2 are presented in Figure 8 \ref{fig:fig8}.
For Experiment 3, we observed differing effects of fine-tuning strategies between Mistral NeMo-based and LLaMA-based models. For Mistral NeMo, directly fine-tuning on Task 2 resulted in better performance than sequential fine-tuning on Task 1 followed by Task 2. Mistral NeMo Base FT-4900 (1) outperformed Mistral NeMo Sequential FT-4900 (Task 1 → Task 2) (2), and Mistral NeMo Base FT-500 (4) similarly outperformed Mistral NeMo Sequential FT-500 (Task 1 → Task 2) (6). However, in the case of LLaMA 3.1-based models, the opposite trend was observed: sequential fine-tuning led to better results. LLaMA Sequential FT-4900 (Task 1 → Task 2) (3) outperformed LLaMA Base FT-4900 (5), and LLaMA Sequential FT-500 (Task 1 → Task 2) (7) surpassed LLaMA Base FT-500 (8). Thus, the effect of sequential fine-tuning varies across models, and no conclusive pattern emerged. The reverse fine-tuning approach, where Mistral NeMo (9) was fine-tuned on Task 2 first and then Task 1, resulted in a notably low METEOR score of 0.16375. Similarly, the Llama-based model (10) showed similar results, with a score of 0.05841. This suggests that the model may have "unlearnt" important aspects of Task 2 during subsequent fine-tuning on Task 1.

\subsubsection{Experiment 4}
\begin{figure}
  \centering
  \includegraphics[width=0.6\textwidth]{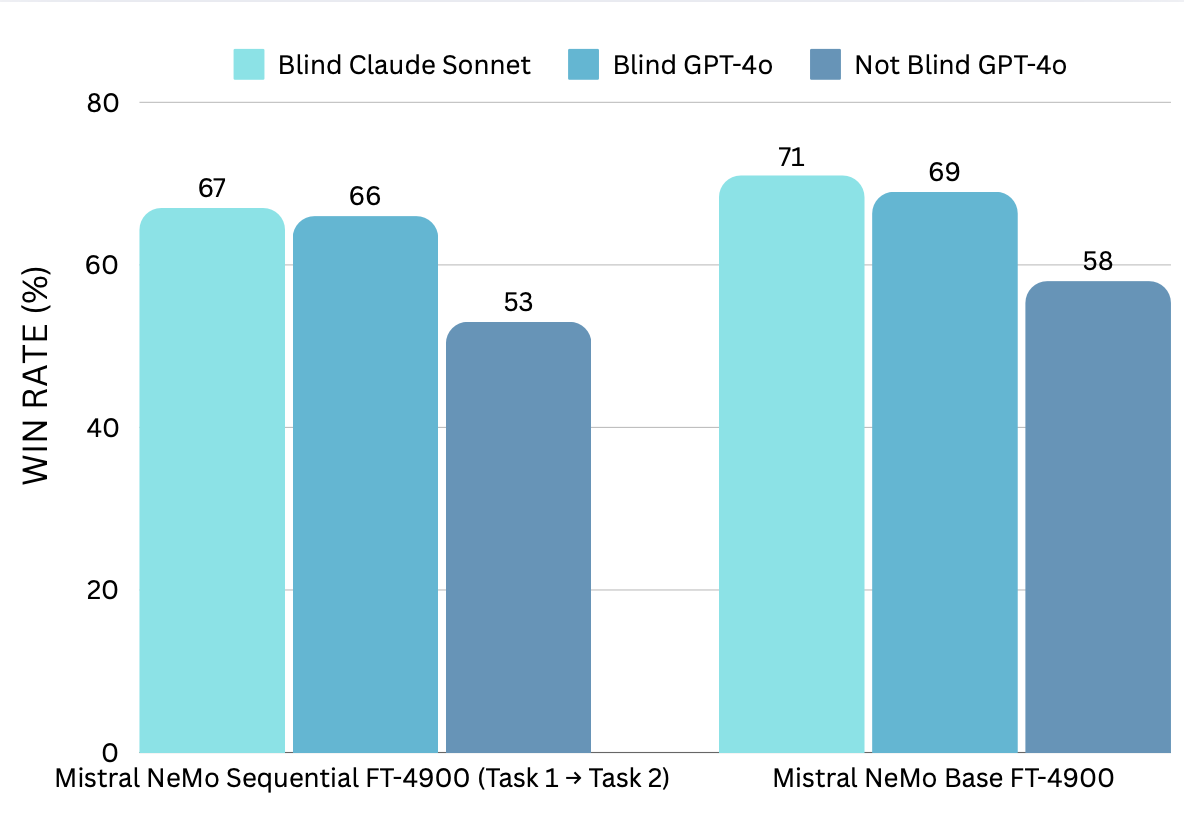}
  \caption{LLM-as-a-judge Win Rate Results}
  \label{fig:fig9}
\end{figure}

For Experiment 4, we selected the two best performing models from Task 2 Experiment 3 above - Mistral NeMo Sequential FT-4900 (Task 1 -> Task 2), Mistral NeMo Base FT-4900 - and compared it against GPT-4 response with LLM-as-a-judge evaluation. The judge LLMs were chosen to be GPT-4o and Claude Sonnet. The first evaluation was conducted without disclosing the identity of the models generating the response. This is marked as the "blind" evaluation. The second evaluation was conducted with the identities revealed. Judgment from Claude Sonnet was used as a benchmark number. Minimal self-recognition bias was detected during the blind GPT-4o evaluation. GPT-4o gave itself only a few more wins than when evaluated by Claude. In the non-blind evaluation, where GPT-4o knew which response it had generated, a significant self-bias appeared, as GPT-4o awarded itself more wins. The actual win-rate dropped by 13 points in non-blind evaluation. This experiment highlights the "narcissistic" bias LLMs tend to have. Figure 9 \ref{fig:fig9} demonstrates the point difference. 

\section{Conclusion, Limitations, and Future Work}
\label{sec:conclusion}
\subsection{Conclusion}
Here are the conclusions from the four experiments conducted, specifically addressing our use cases and the two tasks outlined in this paper:

\textbf{Experiment 1}: Fine-tuning significantly boosts model performance for task-specific applications. Remarkably, these fine-tuned models outperformed GPT-4 while requiring only a fraction of its size and cost. This demonstrates that fine-tuning open-source LLMs is a highly efficient solution for our tasks. Additionally, such models can be hosted privately, addressing critical concerns around patient privacy and data security.

\textbf{Experiment 2}: Expanding the size of the training dataset consistently improved model performance across all tasks, underscoring the importance of data volume in optimizing model accuracy and generalization.

\textbf{Experiment 3}:The effectiveness of sequential fine-tuning strategies varies depending on the model architecture. Furthermore, fine-tuning a model on one task and subsequently on another leads to a significant drop in performance on the first task.

\textbf{Experiment 4}: Non-blind evaluations by GPT-4 resulted in a strong bias, where GPT-4 showed a clear preference for its own outputs. This is consistent with recent research. \cite{liu_llms_2023} noted that LLMs often act as "narcissistic evaluators," inflating their evaluation scores when assessing their own outputs. Similarly, \cite{panickssery_llm_2024} discussed how self-recognition drives this self-preference. However, our study found that this self-recognition bias was more limited unless GPT-4 was explicitly told it was evaluating its own response.

\subsection{Limitations}
One of the key limitations of our work is that we have used synthetic patient data instead of real patient data. This could impact the generalizability of our findings, as synthetic data may not capture the full complexity and variability present in actual patient records.

\subsection{Future Work}
In future work, we plan to investigate multi-task learning (MTL) strategies and continual pretraining (CPT) to improve the performance of models querying Fast Healthcare Interoperability Resources (FHIR) data. By integrating these approaches, we aim to develop a single model capable of efficiently handling both tasks simultaneously.
\hfill

\bibliographystyle{unsrt}  
\bibliography{references}  

\newpage
\appendix

\begin{center}
    \section{Appendix}
\end{center}

\subsection{GPT-4 Prompt for Query Generation: Task 1}
\label{sec:appendix-prompt}

To generate natural language queries, we used the following GPT-4 prompt:

\begin{quote}
\texttt{Pretend you are a patient curious about an aspect of your medical history. Come up with a query that this patient might have regarding their medical data. At least one or more medical data points from the given set of FHIR resources should be sufficient to answer the query. Make the question realistic, simple, and non-technical. For example, 'What are my current medicines?' or 'When was my last shot?' or 'What were the complications of my last heart procedure?;}
\end{quote}
\begin{quote}
\texttt{Generate an output in the JSON format below corresponding to each of the 10 inputted resources after generating 1 query based on one or more of the 10 given FHIR resources: \{10\_resources\}. The relevance should be 'relevant' if the resource was used by the model for the particular query, and 'irrelevant' if not. The resource\_label should be a natural language label generated for each of the 10 resources in the format: 'Condition Cardiac Arrest 06-19-2018'. Therefore, the output should be the 10 JSON formatted files per resource, with patient\_id being the same throughout, relevance can be either relevant or irrelevant if it wasn't used to generate the query. Only one query has to be generated from the 10 resources. So the query label will be the same for all 10. Use the following format for the output:}
\end{quote}

\begin{quote}
\texttt{json [\\
\ \ \ \ \{\\
\ \ \ \ \ \ \ \ "resource": "\{\{resource\}\}",\\
\ \ \ \ \ \ \ \ "query": "\{\{query\}\}",\\
\ \ \ \ \ \ \ \ "relevance": "\{\{relevance\}\}",\\
\ \ \ \ \ \ \ \ "patient\_id": "\{patient\_id\}",\\
\ \ \ \ \ \ \ \ "resource\_label": "\{\{resource\_label\}\}"\\
\ \ \ \ \}\ \ ]\\
}
\end{quote}

\subsection{GPT-4 Prompt for Answer Generation: Task 2}
\label{sec:appendix-prompt3}
\begin{quote}
\texttt{You are a knowledgeable and helpful medical assistant. Answer the given query using the list of relevant FHIR resources provided to you. `Query': \{query\}, `Resources': \{resources\}}
\end{quote}

\end{document}